\documentclass[10pt,twocolumn,letterpaper]{article}

\usepackage{iccv}
\usepackage{times}
\usepackage{epsfig}
\usepackage{graphicx}
\usepackage{amsmath}
\usepackage{amssymb}
\usepackage{stmaryrd}

\usepackage{subcaption}

\usepackage{amsmath}
\usepackage{amssymb}
\usepackage{amsthm}
\usepackage{mathtools}

\DeclareMathOperator{\softmax}{softmax}

\usepackage[pagebackref=true,breaklinks=true,letterpaper=true,colorlinks,bookmarks=false]{hyperref}

\iccvfinalcopy 

\def\iccvPaperID{7911} 

\ificcvfinal\fi

\begin{document}

\title{Motion Guided Attention Fusion to Recognize Interactions from Videos}

\author{Tae Soo Kim  \qquad \qquad Jonathan Jones \qquad \qquad  Gregory D. Hager \\
Johns Hopkins University\\
3400 N. Charles St, Baltimore, MD\\
{\tt\small \{tkim60,jdjones,hager\}@jhu.edu}
}

\maketitle
\ificcvfinal\thispagestyle{empty}\fi

\begin{abstract}
  

We present a dual-pathway approach for recognizing fine-grained interactions from videos.
We build on the success of prior dual-stream approaches, 
but make a distinction between the static and dynamic representations of objects and their interactions explicit by introducing separate motion and object detection pathways. Then, using our new Motion-Guided Attention Fusion module, we fuse the bottom-up features in the motion pathway with features captured from object detections to learn the temporal aspects of an action. We show that our approach can generalize across appearance effectively and recognize actions where an actor interacts with previously unseen objects. We validate our approach using the compositional action recognition task from the Something-Something-v2 dataset where we outperform existing state-of-the-art methods.
We also show that our method can generalize well to real world tasks by showing state-of-the-art performance on recognizing humans assembling various IKEA furniture on the IKEA-ASM dataset.

\end{abstract}

\section{Introduction}
In recent years, “two-stream” approaches have emerged as a dominant paradigm in video-based action recognition \cite{vgg_twostream,Carreira2017QuoVA,feichtenhofer2018slowfast}. Such methods process a video stream using two different neural modules whose scores are fused to produce a final prediction. Each module has its own purpose: typically, one module captures temporal information about motion in the scene, and the other captures spatial information about the appearance of relevant objects, actors, and perhaps background context. 

\begin{figure}[t]
\centering
\includegraphics[width=1.0\linewidth]{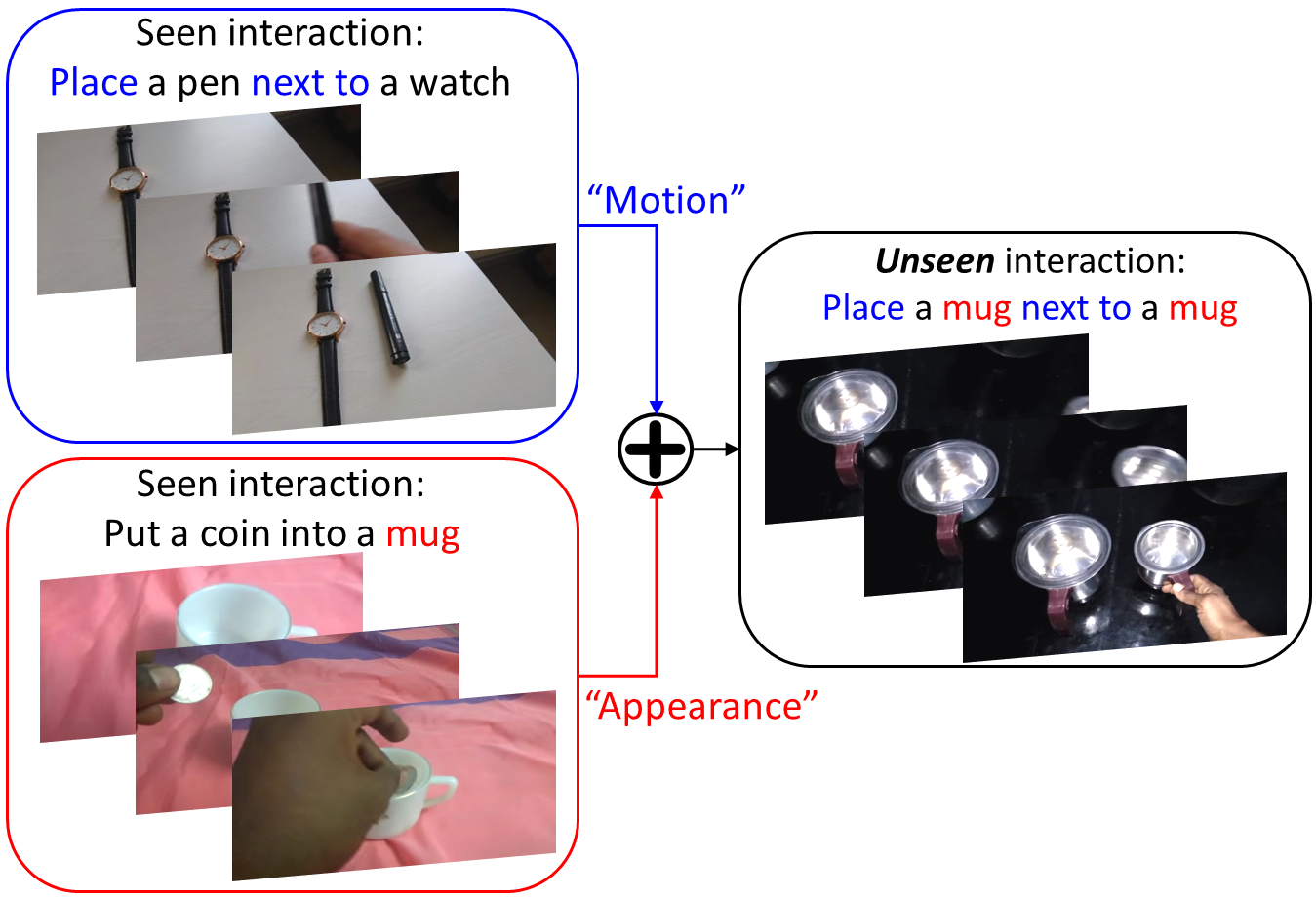}
  \caption{Do current video models have the ability to recognize an unseen instantiation of an interaction defined using a combination of seen components? We show that it is possible by specifying the dynamic structure of an action using a sequence of object detections in a top-down fashion. When the top-down structure is combined with a dual-pathway bottom-up approach, we show that the model can then generalize even to unseen interactions.}
\label{fig:thumbnail}
\end{figure}

Although it is not always explicit in their formulation, two-stream models capture the idea that actions fundamentally describe compositional interactions between people and their environment. These interactions are made up of atomic actions (verbs) which can take a variety of arguments (in analogy to syntactic analyses, subjects or objects). For instance, \texttt{pick up a mug} might be represented as the triple (person, pick up, mug). Due to this compositionality, automated recognition of human object interactions in video thus faces the fundamental challenge that the set of labels is combinatorially large.  As a result, enumerating all possible descriptions to train end-to-end methods \cite{tsm,Carreira2017QuoVA,feichtenhofer2018slowfast,lfb2019,hussein2018timeception} is impractical. As illustrated in Figure \ref{fig:thumbnail}, the compositional nature of interactions ultimately requires a vision system that can generalize to actions with previously seen structure but instantiated with possibly unseen combinations of components.


In response to this challenge, there have been attempts to impose an explicit top-down structure to decompose an action into its actors (objects) and spatial-temporal relation among them using object detections \cite{Wang_videogcnECCV2018,Materzynska_2020_CVPR,Jain2016StructuralRNNDL,Ji_2020_CVPR}.
However, such methods do not exhibit clear and obvious improvements unless ensembled with end-to-end RGB models at test time \cite{Materzynska_2020_CVPR} or ground truth knowledge about actor-object relations is provided to the model by an oracle \cite{Ji_2020_CVPR}. This suggests that enforcing a top-down structure
does not fully capture the range of variations among interactions. Moreover, the fact that an ensemble of independently trained models consistently outperforms an RGB+object feature fusion approach \cite{Materzynska_2020_CVPR} for recognizing interactions suggests that features from the two domains are not effectively fused during training. Nonetheless, the motivation for using object detections  remains strong because objects naturally define a structure of an interaction, and off-the-shelf object detections \cite{maskrcnn,faster_rcnn} have become robust enough to be consumed as-is to define complex actions \cite{daszl}.

In this paper, we present a hybrid approach for recognizing fine-grained interactions that borrows ideas from bottom-up, two-stream action recognition and top-down, structured human-object interaction detection.
The key idea behind our approach is to make use of a sequence of object detections to guide the learning of object-centric video features that capture both static relations and dynamic movement patterns of objects.
The learned object-centric representation is then transferred to a motion pathway using an attention-based  Motion-Guided Attention Fusion (MGAF) mechanism. 
The MGAF module guides the RGB representation from the motion pathway to develop representation of the dynamic aspects of an action. 
In the remainder of the paper, we evaluate the model's ability to generalize over object appearance when recognizing interactions. We show that our approach leads to a video model that can recognize unseen interactions (novel verb-noun compositions) at test time better than existing approaches. We evaluate our approach using the Something-Else task \cite{Materzynska_2020_CVPR} from the Something-Something-V2  dataset \cite{something} where we establish a new state of the art.
Moreover, we show that our framework is a general concept and transfers readily to a new domain. Using the recently released IKEA-ASM dataset \cite{ikea}, we show that our model accurately recognizes humans interacting with numerous parts to assemble various IKEA furniture and also sets the new state-of-the-art benchmark for the main task of the dataset. Further, we present the first results on the compositional task using the IKEA-ASM dataset where a model is tested on novel verb-noun compositions. In summary, the main contributions of the paper are:
\begin{enumerate}
    \item A dual-pathway approach that leverages dynamic relations of objects.
    \item MGAF: A feature fusion strategy to use motion-centric object features to guide the RGB feature learning process.
    \item A state of the art recognition performance on multiple benchmarks including compositional tasks.
\end{enumerate}

\section{Related Work}
\noindent
\textbf{Action Classification in Videos:} With the introduction of large scale video datasets for action classification \cite{caba2015activitynet,ava,kinetics}, many deep bottom-up architectures have been proposed to extract powerful representations from videos \cite{bourdev,Carreira2017QuoVA,feichtenhofer2018slowfast,vgg_twostream,hussein2018timeception,tsm,c3d,lfb2019}. However, the findings in \cite{trn,Materzynska_2020_CVPR} suggest that such pretrained models focus more on appearance rather than the temporal structure of actions. 
We actually build upon this appearance bias to learn useful yet static visual features in the appearance pathway. This representation regarding the static components of an interaction is learned in parallel with the motion pathway which captures the dynamic aspects of the same interaction. We make this explicit by leveraging object detections to guide the motion pathway.

\noindent
\textbf{Top-down Structured Models of Videos with Objects:} A growing line of work uses structured information extracted from videos, such as object detections and scene graphs, to improve fine-grained analysis of actions \cite{Baradel_2018_ECCV,Jain2016StructuralRNNDL,girdhar2019video, lfb2019,Wang_videogcnECCV2018,attendinteract,Sun2018ActorCentricRN,Ji_2020_CVPR,Materzynska_2020_CVPR}. Instead of learning features only from videos, these approaches often combine features extracted from regions of interest defined by object detectors such as \cite{faster_rcnn,maskrcnn}. The object centric representations can then be used to learn pairwise relations between objects \cite{Materzynska_2020_CVPR,attendinteract,Baradel_2018_ECCV}, between objects and global context \cite{lfb2019,Sun2018ActorCentricRN,girdhar2019video,Materzynska_2020_CVPR} and within a specified graph structure \cite{Wang_videogcnECCV2018,Ji_2020_CVPR} to improve action classification. We also use object detections to provide the structure of interactions. However, we take a more data-driven approach to learn the structure from a sequence of objects instead of specifying them in a purely top-down fashion. Our approach then uses the learned object-centric concepts to guide the motion pathway to learn more motion-centric features from videos.

\noindent
\textbf{Human Object Interactions:} Detecting human-object interaction (HOI) from a still image is an active area of research \cite{hico,hico-det,GuptaM15,tamura2020bcar}. Please see the very recent state-of-the-art HOI paper \cite{tamura2020bcar} for more review of the image-based HOI literature. We fundamentally differ from image-based method in that we process a sequence of images and focus on modeling the dynamic aspects of interactions.

In the video domain, authors of \cite{interaction-hotspots} define `interaction hotspots' and learn object affordances from videos. Though the method was not used to recognize actions in a video, it provided evidence that the model can correctly compute affordances of objects that did not appear in the training set. Earlier works for recognizing interactions also exploited the top-down structure provided by object detections. However, given the hand-designed nature of the pair-wise object attributes, manually specified methods such as \cite{Escorcia_2013_ICCV_Workshops} do not scale well. Authors of \cite{Zhou_2015_CVPR} propose an extension where the structure is specified implicitly by extracting descriptors along object tracks. However, we later show in our experiments that combining object level movement information with static visual information leads to best results. Our contribution includes how we merge the information between the two modalities using our attention-based MGAF module.

\noindent
\textbf{Use of Attention: }
Recent papers have investigated how the self-attention formulation from the natural language processing domain \cite{Vaswani_attention,brown2020language,devlin-etal-2019-bert,roberta,NIPS2019_8812} can generalize to the image domain \cite{AAconv,standalone,Cordonnier2020On} as well as to video-based applications \cite{NL_2018_CVPR}. The Non-local Neural Network \cite{NL_2018_CVPR} captures long-range dependencies within a video by use of a non-local operator which is a generalization of the self-attention unit \cite{Vaswani_attention}. Our key observation is that the attention operation leads to a fusion of information whether it is within the model's spatial-temporal feature maps \cite{NL_2018_CVPR} or between long-term and short-term features \cite{lfb2019} in a feature-bank framework. We investigate whether the attention mechanism can be used to guide the RGB representation to focus more on the dynamic aspects of an action by fusing information from the object features.


\noindent
\textbf{Architecture:} The well known two-stream architecture \cite{vgg_twostream} has been modified and adopted in state-of-the-art video recognition models such as \cite{Carreira2017QuoVA,feichtenhofer2018slowfast}. The key idea of the original two-stream method is that the stream that takes as input optical-flow fields learns features regarding motion. Our approach to guide the Motion-pathway representation with object features shares the same philosophy. However, we are imposing a top-down structure that is more relevant to modeling the relation of objects within an interaction. The more modern SlowFast \cite{feichtenhofer2018slowfast} architecture also uses two RGB streams to learn features with different temporal granularities by processing the video at different temporal sampling rates for each stream. Instead, we use the same framework to explicitly dedicate a pathway to learning features related to dynamic and static aspects of an action. We make this separation more explicit by leveraging temporal features extracted from object detections.



\begin{figure*}[t]
\centering
\includegraphics[width=0.9\linewidth]{ 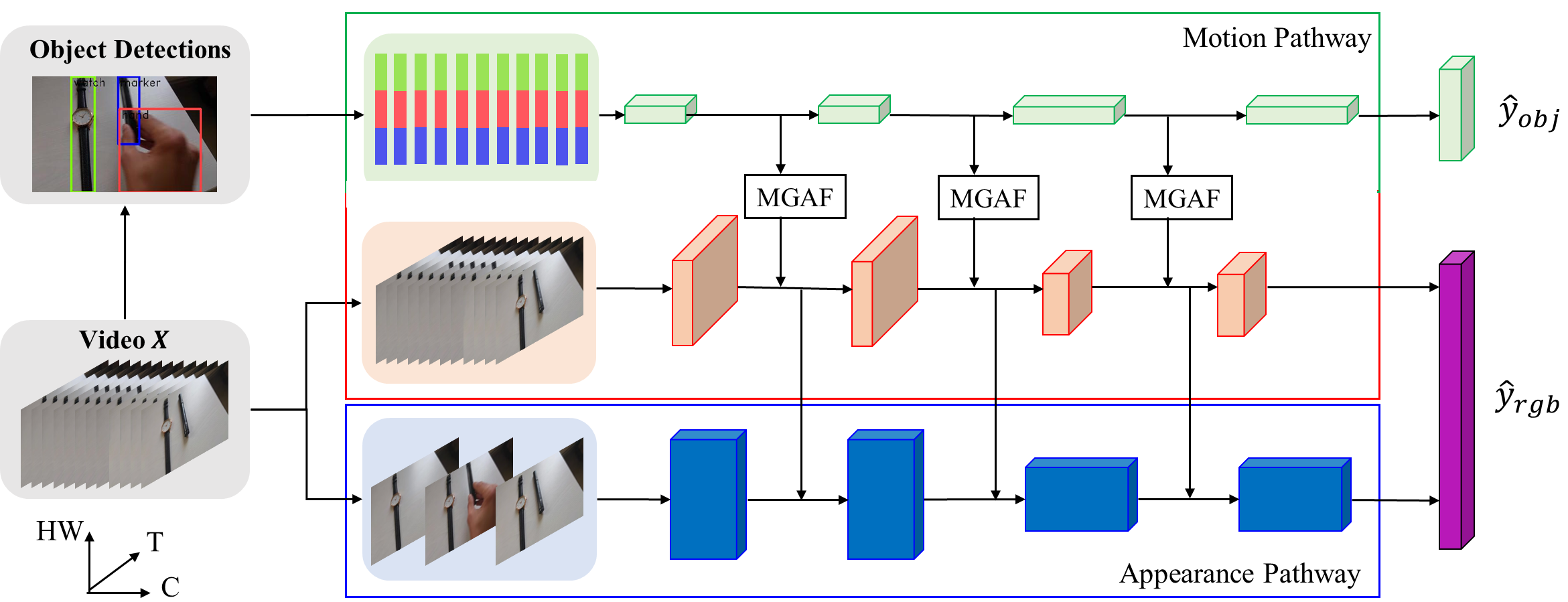}
  \caption{Our approach processes a video using two pathways that capture different aspects of an interaction. The appearance pathway learns static visual cues from a video using using only a few frames sampled from a video. The motion pathway explicitly captures dynamic information of the action from a video by leveraging the temporal features extracted from object detections. The Motion Guided Attention Fusion (MGAF) module effectively fuses the top-down structural inforamtion provided by the object detections to guide the representation learning process of the motion pathway. }
\label{fig:architecture}
\end{figure*}

\section{Method}
We describe the components of our dual-pathway for recognizing interactions from videos. The two pathways, Appearance (Sec \ref{sec:appearance}) and Motion (Sec \ref{sec:motion}), both take as input a RGB video but only the Motion pathway fuses information using object detections. We describe how we learn high-level motion cues using object detections in (Sec \ref{sec:objects}) with a simple temporal model. Finally, we introduce the Motion-Guided Attention Fusion (MGAF) module (Sec \ref{sec:mgaf}) that fuses features from the Motion pathway with the object-centric features using a multi-modal attention operation. 

\subsection{Appearance pathway: learning static content}
\label{sec:appearance}
We exploit the appearance bias \cite{Materzynska_2020_CVPR, trn} of modern 3D convolutional models such as \cite{Carreira2017QuoVA} to our advantage and use them to construct the Appearance pathway. 

Let $X \in \mathbb{R}^{T \times H \times W \times C}$ be a video with $C$ channels with $T$ frames with spatial dimension $H$ and $W$. The Appearance pathway can be any feed-forward neural architecture $V$ that has the following form:
\begin{equation}
V(X) = v_L(v_{L-1}(\dots v_2(v_1(X)))) 
\end{equation}
where the $l$-th intermediate representation is computed by sub-modules $v_{1:l}$ of the network and $V(X) \in \mathbb{R}^{T_V^L \times H_V^L \times W_V^L \times C_V^L }$. There are many well-established convolutional neural architectures that satisfy the above conditions \cite{Carreira2017QuoVA,tsm,feichtenhofer2018slowfast,vgg_twostream,r21d} and our general framework supports the use of any such models. 

Since the goal of the Appearance pathway is to capture the static components of an action, we use a low sampling rate to sub-sample the video. Conveniently, most aforementioned 3D convolution architectures \cite{Carreira2017QuoVA,tsm, feichtenhofer2018slowfast,p3d,x3d,r21d} already follow such a video sampling strategy.

\subsection{Learning high-level motion features using object detections    }
\label{sec:objects}
The main insight of our approach is that  objects over time encode the characteristic movement patterns of interactions. 
Suppose we can detect at most $D$ objects in a given video with $T$ frames. Let $Z(X) \in \mathbb{R}^{T \times 4D}$ be a representation of the video $X$ defined using object detections. We define a frame $Z(X,t) \in \mathbb{R}^{4D}$ as a concatenation of $D$ bounding box coordinates of detected objects such that:
\begin{equation}
    Z(X,t) = [o^1_{x1},o^1_{y1},o^1_{x2},o^1_{y2},\dots,o^D_{x1},o^D_{y1},o^D_{x2},o^D_{y2} ]
\end{equation}
where $o^d_{x1},o^d_{y1},o^d_{x2},o^d_{y2}$ corresponds to the bounding box coordinates of the $d$-th object category. In practice, we set $D$ as a constant and zero-pad the appropriate dimensions when there are few than $D$ objects in the scene. When there are more than $D$ objects, we select $D$ objects based on their prediction confidence scores. 

Given a time-series of frame-level object detections, let $U$ be a feed-forward architecture such that:
\begin{equation}
U(X) = u_L(u_{L-1}(\dots u_2(u_1(Z(X))))) 
\end{equation}
where $u_l$ is a submodule of $U$, $1 \leq l \leq L$  and $U(X) \in \mathbb{R}^{T_U^L \times  C_U^L}$. Our framework does not require the depth of the $V$ and $U$ to be the same; it only requires that $U$ contains a sequence of trainable layers that performs temporal feature extraction. For example, each $u_l$ can be implemented as a temporal convolution layer with 1D convolutions followed by a non-linear operation \cite{Lea2017TemporalCN}, a recurrent layer such as \cite{lstm,gru}, a self-attention based transformer encoder \cite{Vaswani_attention} or any mixture of such components. 


When we optimize $U$ to predict interaction labels from time-series of object detections, $u_l$ by design contains information regarding relations between objects and their dynamics over time. The key aspect of our approach is to transfer the object-centric features learned from $u_{1:L}$
to the RGB-based Motion pathway which we describe next.

\subsection{Motion pathway: learning dynamic structure from objects and video}
\label{sec:motion}

The Motion pathway assumes the same input video $X \in \mathbb{R}^{T \times H \times W \times C}$ as the Appearance pathway. The goal of the Motion pathway is to extract motion-biased features from $X$ by learning to fuse the object-motion features provided by $U$. Suppose the Motion pathway $M$ is a feed-forward architecture with $L$ modules $m_{1:L}$. Given the output of the previous layer $M_{l-1} \in \mathbb{R}^{T_M^{l-1} \times H_M^{l-1} \times W_M^{l-1} \times C_M^{l-1} }$, each module $m_l$ is defined as a residual block \cite{r21d} with a temporal convolution operation followed by a spatial convolution:
\begin{equation}
\begin{split}
    f_l &= \sigma(F(M_{l-1};\theta^f_l))  \\
    g_l &= \sigma(G(f_l;\theta^g_l))\\
    m_l(M_{l-1}) &= M_{l-1} + g_l
\end{split}
\end{equation}
where $F$ is a temporal 3D convolution operation where each filter in $\theta^f_l$ has a size $t \times 1 \times 1 \times C_M^{l-1}$, $G$ is a spatial 3D convolution operation where each filter in $\theta^g_l$ has a size $1 \times k \times k \times C_M^{l-1}$, $\sigma$ is a normalization operation followed by a non-linearity, and $t,k$ are temporal/spatial filter dimensions. 

\begin{figure}[t]
  \subfloat[RGB-only]{
	\hfill
	\begin{minipage}[c][]{
	   0.17\textwidth}
	   \centering
	   \includegraphics[width=1\textwidth]{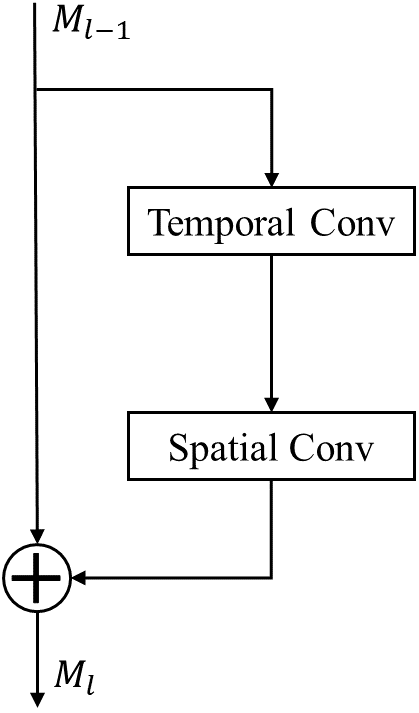}
	\end{minipage}}
  \subfloat[RGB+objects with MGAF]{
	\begin{minipage}[c][]{
	   0.29\textwidth}
	   \centering
	   \includegraphics[width=1\textwidth]{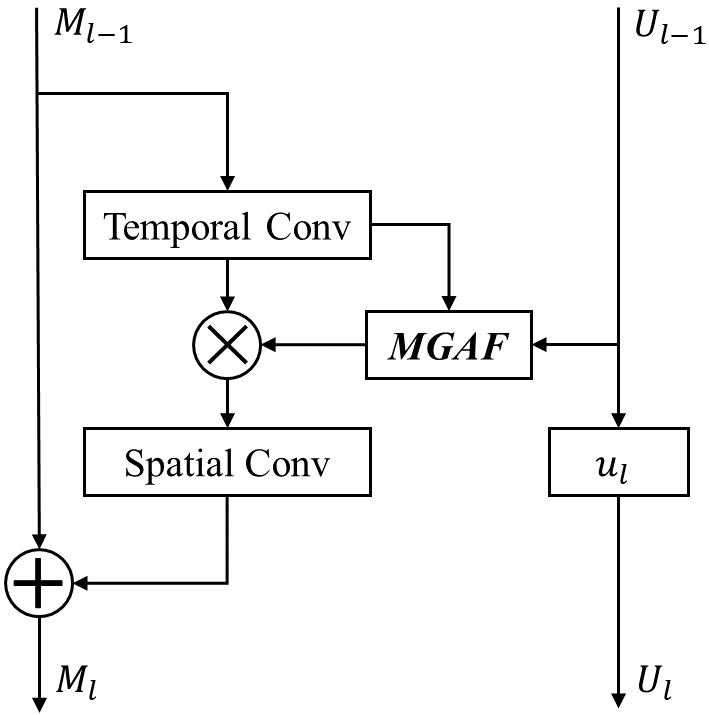}
	\end{minipage}}
	 \hfill 	
\caption{ (a): An illustration of the $l$-th module in the motion pathway that learns features over a visual feature $M_{l-1}$ to produce $M_l$. (b) The same module augmented with a Motion Guided Attention Fusion (MGAF) which fuses the RGB feature $M_{l-1}$ with the object feature $U_{l-1}$ to yield a more motion-centric representation. }
\label{fig:modules}
\end{figure}

Suppose the module has access to motion features $U_{l-1} \in \mathbb{R}^{T_M^{l-1} \times C_U^{l-1} } $ as an additional input that has the same temporal length and some per-frame feature dimension $C_U^{l-1}$. We later show in our experiments that \textit{where} and \textit{how} $U_{l-1}$ gets fused with the Motion pathway module is critical. To best preserve temporal information when fusing, we choose the representation $f_l$ resulting from a set of temporal convolutions to fuse with $U_{l-1}$. To learn to merge only the relevant motion information, we introduce the Motion-Guided Attention Fusion (MGAF) to fuse the visual representation $f_l$ with the object feature $U_{l-1}$. We modify the module $m_l$ accordingly as:
\begin{equation}
\begin{split}
    f_l &= \sigma(F(M_{l-1};\theta^f_l))  \\
    fused_l &= \text{MGAF}(f_l,U_{l-1})  \\
    g_l &= \sigma(G(fused_l;\theta^g_l))\\
    m_l(M_{l-1},u_{l-1}) &= M_{l-1} + g_l
\end{split}
\end{equation}
Figure \ref{fig:modules} visualizes the operation within a block in the motion pathway.
We describe how we perform multi-modal feature fusion using the MGAF module.

\subsection{MGAF: Motion Guided Attention Fusion}
\label{sec:mgaf}
A common feature fusion strategy is to concatenate the two representations along the channel dimension. Concatenation assumes that all channels of the two features contribute equally. Instead, we would like to enhance only the channels that capture relevant motion patterns. For this purpose, we allow the the RGB features $f_l$ to attend to the object-centric representation $U_{l-1}$  and effectively re-calibrate the channels of the $f_l$ via a cross-modal attention operation.

Given $f_l \in  \mathbb{R}^{T_M^{l} \times H_M^{l} \times W_M^{l} \times C_M^{l}}$, we first perform spatial pooling with a window of size $H_M^{l} \times W_M^{l}$ such that $\text{pool}(f_l) = z(f_l) \in  \mathbb{R}^{T_M^{l}\times C_M^{l}}$. For notational brevity, we drop subscripts and represent $z(f_l)$ as $z$ and $U_{l-1}$ as $U$. Then, we allow the spatially collapsed visual representation $z$ to attend to the object feature $U$ by:
\begin{equation}
\label{eq:sif1}
A_{z \shortrightarrow U}  =  \softmax( \frac{(zW_z)(W^T_UU^T)}{\sqrt{C}})UW_U
\end{equation}
where $W_z \in \mathbb{R}^{C_M^l \times C}$ and $W_U \in \mathbb{R}^{C_U^{l-1} \times C}$ are learnable parameters of the MGAF module and $C$ is a hyperparameter. Then the attention $A_{z \shortrightarrow U}$ feature is used to re-weight channels of $z$ by:
\begin{equation}
    \text{MGAF}(f_l,U_{l-1}) = \sigma(\alpha(A_{z \shortrightarrow U})W_{uz}) \otimes f_l
\end{equation}
where $\alpha$ is a normalization (layer-norm) operation followed by an activation operation, $W_{uz} \in \mathbb{R}^{C \times C_M^l}$ is a learnable transformation, $\sigma$ is a sigmoid function and $\otimes $ is an element-wise multiplication. The term $\sigma(\alpha(A_{S \shortrightarrow V})W_{uz})$ acts as a gating mechanism to re-calibrate both the channel and time dimensions of $f_l$ based on the attention operation between the RGB and object features.


\begin{figure}[t]
	   \centering
	   \includegraphics[width=1\linewidth]{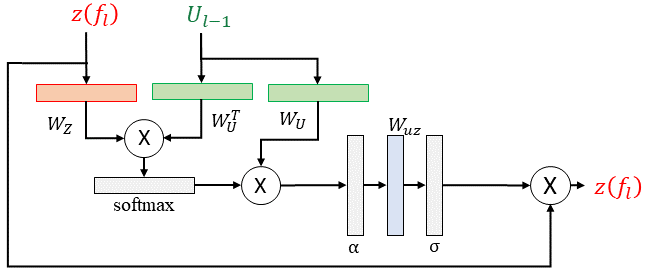}
\caption{The Motion Guided Attention Fusion module. We fuse the information between the spatially collapsed RGB feature $z(f_l)$ from the motion pathway and the object feature $U_{l-1}$. We use a self-attention mechanism to achieve this multi-modal feature fusion.} 
\label{fig:mgaf}
\end{figure}

\subsection{Instantiation}
Figure \ref{fig:architecture} shows the overall architecture of our approach. We use the two stream architecture inspired by \cite{feichtenhofer2018slowfast}. Likewise, we use separate frame rates for each pathway: the Appearance pathway processes a video with a very low frame rate and the Motion pathway extracts visual features from a video with a higher frame rate (higher by a factor of $\alpha=8$). We later show in our experiments that the difference in frame rate alone does not necessarily lead to decoupled representation of motion and appearance which limits the model's generalization performance. We show that the fusion of features derived from time-series of object detections is necessary for the model to explicitly learn representation regarding the motion patterns of interactions. 

As illustrated in Figure \ref{fig:architecture}, we keep the relative feature dimensions consistent with the original formulation in \cite{feichtenhofer2018slowfast} where the Motion pathway has fewer convolutional kernels by a factor of $\beta=1/8$. We also do not change the fusion strategy (conv-fusion) of information between the two RGB pathways as in \cite{feichtenhofer2018slowfast} to keep comparisons simple and fair. Finally, we learn to classify interactions by finding the optimal set of trainable parameters for all modules jointly by minimizing the following loss:
\begin{equation}
    \mathcal{L} = -\sum_N (\lambda_ry^n\text{log}(\hat{y}^n_{r})+\lambda_oy^n\text{log}(\hat{y}^n_{o})) 
\end{equation}
where $y^n$ is the ground truth interaction label of the $n$-th training sample, $\hat{y}^n_{r}$ is the prediction using the RGB feature, $\hat{y}^n_{o}$ is the prediction with the object features, and $\lambda_r,\lambda_o$ are parameters to control to contribution of each cross-entropy terms.


\section{Experiments on Something Else}
\label{sec:ss_results}

Labels found in the Something-Something-V2 \cite{something} dataset have a compositional structure where a combination of a verb and a noun (object) defines an action. The dataset contains a total of 174 action categories where a crowd-sourced worker uploads a video capturing an arbitrary composition of an action category (verb) with an object (noun). As a result, the data set contains a very diverse set of verb-noun compositions involving 12,554 different object descriptions. The recently released Something-Else task \cite{Materzynska_2020_CVPR} is an extension to the original task with new object annotations and a compositional action recognition task. 

\subsection{The compositional action classification task}
The new compositional split assumes that the set of verb-noun pairs available for training is disjoint from the set given at inference time.

Let there be two disjoint sets of nouns (objects), $\{ \mathcal{A}, \mathcal{B} \}$, and two disjoint sets of verbs (actions) $\{1,2\}$. The goal of the compositional action recognition task is to recognize novel verb-noun compositions at test time. The model can observe instances from the set $\{1\mathcal{A}+2\mathcal{B}\}$ during training but will be tested using instances from $\{1\mathcal{B}+2\mathcal{A}\}$. In this setting, there are 174 action categories with 54,919 training and 57,876 validation instances. The model is evaluated using a standard classification set up and we measure performance with top-1 and top-5 accuracies.

\subsection{Implementation detail}
The presented framework is general and can use most recent state of the art models to instantiate each components. We extend the SlowFast \cite{feichtenhofer2018slowfast} architecture given its dual-pathway implementation and state of the art results on large-scale action classification benchmarks. We adopt the Slow pathway from \cite{feichtenhofer2018slowfast} as our appearance and the Fast pathway as the motion pathway. The appearance and motion pathways subsample 8 and 32 frames respectively given a video sample. 

We use the ground truth object detections and tracks provided by the dataset release. For results using predicted object detections, we use the same detection boxes as the authors of \cite{Materzynska_2020_CVPR} which come from a trained Faster-RCNN \cite{faster_rcnn} with the Feature Pyramid Network (FPN) \cite{fpn} and ResNet-101 \cite{resnet} backbone. The object detector outputs a set of a person (hand) and generic-object localizations as well as confidence scores. 

In terms of the object-based temporal model, we use a very light-weight 5-layer temporal convolutional neural network \cite{tcn,restcn}. We do not perform any pooling operation in the temporal dimension until the final global-average pooling layer. All temporal convolution filters are of length 9 with stride 1. All experiments are performed using the Pytorch \cite{pytorch} framework. Additional details necessary to reproduce our results including optimization settings, hardware specs and training parameters are reported in the supplementary material.


\begin{table}[t]
\centering
\begin{tabular}{l|cc|cc}
                \multicolumn{1}{c|}{Model} & \multicolumn{2}{c|}{Input} & \multicolumn{2}{c}{Evaluation} \\ \hline
 & RGB         & Objects         & top-1      & top-5\\ \hline
SP* -- mixed \cite{Carreira2017QuoVA} & o  & & 61.7 & 83.5  \\
DP* -- mixed \cite{feichtenhofer2018slowfast} & o&  & 64.9 & 90.1  \\
STIN* -- mixed \cite{Materzynska_2020_CVPR} &  & o  & 54.0 & 79.6  \\
Ours* -- mixed &  & o  & 55.1 & 79.9  \\\hline \hline
SP \cite{Carreira2017QuoVA} & o  & & 46.8 & 72.2  \\
DP \cite{feichtenhofer2018slowfast} & o & &49.6&77.9  \\\hline
STIN \cite{Materzynska_2020_CVPR}& & o & 51.4&79.3\\
STIN -- concat & o & o & 54.6 &79.4 \\
STIN -- ensemble    & o & o & 58.1 &83.2\\\hline
Ours (Obj only) &  & o & 52.3 & 77.5 \\
MGAF(SP, Obj)  & o & o & 60.5 & 84.3\\
MGAF(DP, Obj)  & o & o & \textbf{68.0} &  \textbf{88.7} \\
      
\end{tabular}
\caption{Comparison to other methods on the compositional action classification task using ground-truth objects. SP: Single-pathway. DP: Dual-pathway. MGAF: Motion-guided attention fusion. *--mixed : Indicates that the verb-noun compositions found during training also exist in the test set.}
\label{tab:ss_comp}
\end{table}

\subsection{Comparison to the state of the art} For RGB-only baselines, we use the popular I3D \cite{Carreira2017QuoVA} model as our single-pathway (SP) architecture and the SlowFast \cite{feichtenhofer2018slowfast} model as the dual-pathway (DP) baseline. In table \ref{tab:ss_comp}, we first evaluate the performance of the models when the set of verb-noun compositions available during training is not disjoint (*--mixed) from the set found during testing. We observe a large drop in performance of about  $25\%$ for both SP and DP models. This is a strong evidence that current methods for bottom-up video classification do not generalize well across different verb-noun compositions. The purely bottom-up models are possibly too biased towards appearance of actions and fail to generalize across object appearance involved in the interactions. 

Using the top-down structure provided by object detections, we show in Table \ref{tab:ss_comp} that models that only use object detections (such STIN \cite{Materzynska_2020_CVPR} and our object-only temporal model) already outperform purely bottom-up RGB-only models. This shows that object detections provide strong structural cues necessary for recognizing interactions.

The key differentiating aspect of our approach compared to other state of the art models such as the STIN \cite{Materzynska_2020_CVPR} is how we leverage the top-down structure extracted from object detections to guide the bottom-up feature learning process from videos. Table \ref{tab:ss_comp} provides a head to head comparison of the state-of-the-art model (STIN) to our approach. To learn a joint model of video and objects, STIN concatenates object-based features to visual features extracted from an I3D model. The resulting STIN--concat model improves over the object-only STIN baseline by 2.8 points. In comparison, the MGAF(SP,Obj) model uses the MGAF module instead to fuse features from object detections and the same I3D model. We observe a significantly larger gain of 8.2 points over our object-only baseline model. This actually outperforms even the ensemble of multi-modal approaches (STIN--ensemble).

The MGAF(SP,Obj) instantiation still learns visual representation from video using only a single pathway. We can make the decoupling of the motion and appearance representations more explicit by using the dual-pathway formulation. As described in the Methods section, we fuse only the motion pathway with the object based features. The resulting MGAF(DP,Obj) model leads to significant improvements in performance, leading to a state-of-the-art performance of 68.0 top-1 accuracy. 

When using predicted object detections instead of ground truth localizations, we observe that the noise in object locations causes the performance of MGAF(DP,Obj) to drop to top-1 and top-5 accuracies of 61.2 and 83.3. This is still an improvement of 9.3 and 6.2 points over the current state of the art \cite{Materzynska_2020_CVPR} which performs at 51.5 top-1 and 77.1 top-5 accuracies.

\begin{table}[t]
  
    \begin{subtable}[h]{0.45\textwidth}
        \centering
        \begin{tabular}{c|cc}
        Obj-only Model & Comp. &\# Params \\ \hline
        STIN \cite{Materzynska_2020_CVPR} & 51.4 &    4.288M  \\
        v1 & 52.3 &  0.838M  \\
        v2 & 53.6 &  4.150M     
        \end{tabular}
       \caption{Variantions of our object-only model compared to STIN.}
       \label{tab:ablation2}
    \end{subtable}
    \hfill
        \vspace{0.5cm}
        
    \begin{subtable}[h]{0.45\textwidth}
        \centering
        \begin{tabular}{c|c}
  & Top-1 Acc.\\\hline 
A only & 46.8 \\
M only & 39.4 \\
A + M (Dual-pathway) & 49.6 \\ \hline
O only & 52.3 \\
Concat(A, O) & 54.7 \\
Concat(A + M , O) & 58.8 \\ \hline
MGAF(A, O) & 60.5 \\
MGAF(M, O) & 55.8 \\
M + MGAF(A, O) & 63.8 \\
A + MGAF(M, O) & \textbf{68.0} \\
\end{tabular}
        \caption{Ablations on model components using the Something-Else compositional task. \textbf{First block}: comparison of RGB-only components. \textbf{Second block}: naive concatenation approach to fuse object features (O) with RGB features (A and M).  \textbf{Third-block}: Comparisons of different input combinations to the MGAF module. A: Appearance-pathway. M: Motion-pathway. O:Object-pathway}
        \label{tab:ablation1}
     \end{subtable}
         \hfill
         \vspace{0.50cm}
         

    
     \caption{Various ablations of model components using the compositional split of the Something-Else dataset.}
     \label{tab:ablations}
\end{table}
\subsection{Ablations}
In Table \ref{tab:ablation2}, we compare variations of our simple temporal model that learns from a sequence of object detections. We wanted it to be as fast and as light-weight as possible such that it adds minimal overhead to the video model. The v1 and v2 object models both have the same depth (5 temporal convolution layers) but with different number of filters per layer. We use the v1 model throughout all our Something-Else experiments because it still outperforms the state of the art STIN with less than $20\%$ of learnable parameters than either STIN or our v2 variant. 

In Table \ref{tab:ablation1}, the first block of rows compares the contribution of each pathway of the DP given only RGB videos. We show that the combination of both the appearance (A) and motion (M) pathways is necessary to improve recognition performance of interactions found in the Something-Else dataset. The second block of Table \ref{tab:ablation1} shows the performance of a model that combines RGB and object detections via a concatenation of features from the two domains. The Concat(A,O) is essentially a late-fusion of multi-modal features from the SP and object (O) models. We observe that the top-down structure provided by O helps improve the model over the object-only baseline by 2.4 points and over the RGB-only baseline by 7.9 points. We find consistent behavior when the object feature O is concatenated with the output of the RGB only dual-pathway (A+M) model.

Compared to the Concat(A,O) model, the comparable model using the MGAF module, MGAF(A,O), achieves a considerably higher accuracy (54.7 vs. 60.5). This fuses the appearance feature directly with the object features but without the use of a separate motion pathway. Given that there are two RGB pathways (A and M), the MGAF module can be used to fuse the object-centric representation O with either pathway. We find that fusing O with the motion pathway M and then later merging with A leads to best results (68.0). We believe the finer temporal granularity of the motion pathway input preserves more dynamic information of the video and thus fuses more effectively with O.

\section{Experiments on the IKEA Assembly dataset}
\label{sec:ikea}

\begin{figure}[t]
\centering
\includegraphics[width=1.0\linewidth]{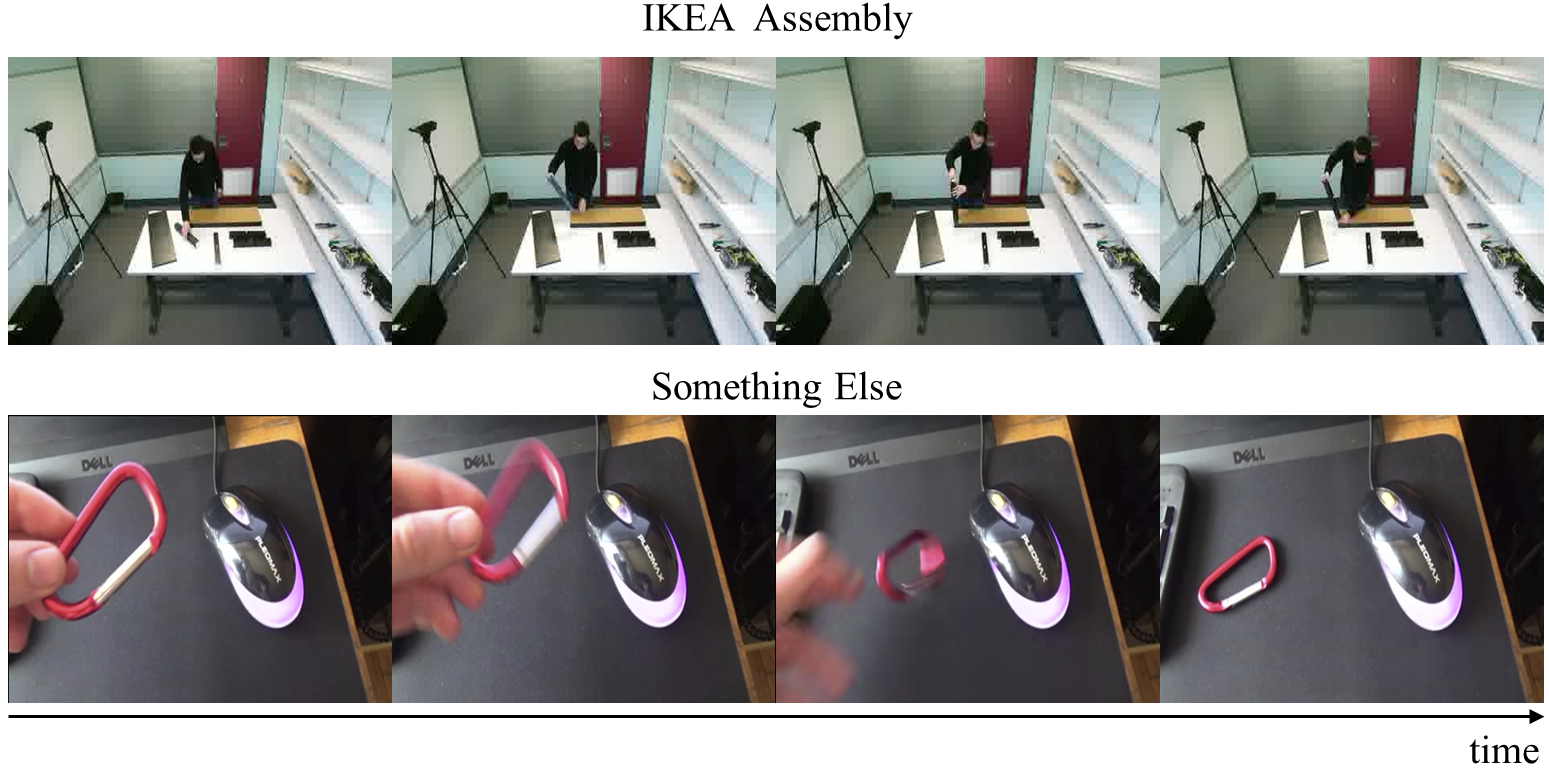}
  \caption{The difference between the IKEA-Assembly and the Something-Else datasets include scale, granularity of motion, viewpoint and levels of occlusion.}
\label{fig:compare}
\end{figure}

In this section, we test the ability of our model to recognize realistic human object interactions using the IKEA Assembly dataset. Compared to the Something-Else dataset, the IKEA dataset contains realistic interactions that occur at a much more granular scale. Figure \ref{fig:compare} clearly illustrates the differences in viewpoint, object scale and levels of occlusion between the two datasets. We show that our approach transfers well to this realistic domain. 

A label in the IKEA-Assembly dataset is defined by a composition of a verb (ie. spin) and an object (ie. a leg). There are 12 verbs and 7 objects present in the dataset. This leads to a total of 33 defined interactions. 
The compositional structure of interactions gives rise to a severely unbalanced label set. For example, there are 754 training examples of \texttt{spin leg} as opposed to only 20 samples of \texttt{lay-down leg}. Hence, 
we report both the micro averaged accuracy and mean of per class recall (macro-recall) to assess the models. The implementation detail necessary for reproducing our results will be detailed in the supplementary material.

\begin{table}[h]
\centering
\begin{tabular}{l|cc|cc}
                \multicolumn{1}{c|}{Model} & \multicolumn{2}{c|}{Modality} & \multicolumn{2}{c}{Evaluation} \\ \hline
 & RGB         & Objects         & Macro  & Micro\\ \hline
SP (I3D \cite{Carreira2017QuoVA}) & o  & & 41.8 & 74.6  \\
DP (SlowFast \cite{feichtenhofer2018slowfast})  & o  & & 43.9 & 73.5\\
Obj-only  &  & o & 18.9& 57.8 \\ \hline
Concat(SP, Obj) & o  & o & 44.2 & 76.2 \\
Concat(DP, Obj)  & o  & o  & 46.0 & 76.5 \\
MGAF(DP, Obj)  & o  & o & \textbf{47.7} & \textbf{78.8 }\\
\end{tabular}
\caption{Results on the original task of the IKEA-Assembly dataset. SP:Single-pathway DP: Dual-pathway. Concat: conatenation of features. MGAF: Motion Guided Attention Fusion}
\label{tab:ikea_original}
\end{table}
\begin{table}[t]
\centering
\begin{tabular}{l|cc|cc}
                \multicolumn{1}{c|}{Model} & \multicolumn{2}{c|}{Mixed} & \multicolumn{2}{c}{Compositional} \\ \hline
 & Macro & Micro & Macro  & Micro\\ \hline
SP (I3D \cite{Carreira2017QuoVA}) & 44.8 & 66.4 &  27.0& 45.1\\
Obj-only  & 24.7 & 37.4 & 22.4 & 42.1  \\ 
Concat(SP, Obj) & 45.6 & 68.7 &  28.3 & 43.1  \\ \hline
DP (SlowFast \cite{feichtenhofer2018slowfast})  & 48.8  & 72.9 & 29.4  & 54.7 \\
Concat(DP, Obj)  & 49.0 & 73.2  & 32.0 & 53.7  \\
MGAF(DP, Obj)  & 49.1 & 72.4 & 37.6 & \textbf{55.6}  \\
\end{tabular}
\caption{Results on the compositional task of the IKEA-Assembly dataset. SP:Single-pathway DP: Dual-pathway. Concat: concatenation of features. MGAF: Motion Guided Attention Fusion}
\label{tab:ikea_comp}
\end{table}
\subsection{Results on the original task}
We evaluate our approach on the original task of the dataset. The verb-noun compositions available during training also appears at test time in this setup. This is equivalent to the `mixed' compositional setup shown in Section \ref{sec:ss_results}. In Table \ref{tab:ikea_original}, we first report the performances of RGB-only baselines, the single-pathway I3D \cite{Carreira2017QuoVA}) and the dual-pathway SlowFast \cite{feichtenhofer2018slowfast}. In this dataset, we find that the there is no significant performance gap between the SP and DP models. This suggests that the extra motion pathway in DP is not contributing much. We believe this is caused by a combination of two factors. First, IKEA-Assembly dataset is much smaller than the Something-Else dataset (around 5k training instances vs. 50k) hence the purely bottom-up DP model might not have fully learned to decouple motion and appearance. Second, given the experimental setup, many interactions can be correctly classified using just the static cues (ie. \texttt{lay-down leg} vs. \texttt{pick-up shelf}).

We find that the top-down structure given by the object detections helps mitigate the first issue. For instance, Concat(DP, Obj) instantiation improves the RGB-only DP baseline by 2.1. We gain an additional 1.7 points when using the MGAF module to target the fusion to the motion pathway. However, as mentioned above, the model can get away with not having to model motion explicitly in this setup. Next, we describe the compositional task where a model must be able to explicitly reason about the dynamic as well as the static components of an interaction.

\subsection{Results on the compositional task}
We introduce the compositional task for the IKEA-Assembly dataset. The setup here is the same as the compositional task from the Something-Else dataset. In essence, we are testing the model's ability to recognize a `\textit{push table top}' instance that it has not seen during training by observing `\textit{push} leg' and `flip \textit{table top}' samples during training. This leads to a 6-way classification of verbs. We provide the details of how we split the action labels to form our compositional split in the supplementary material.

In Table \ref{tab:ikea_comp}, we observe that RGB-only models (SP and DP) show large discrepancies in performance between the mixed and compositional tasks. The big performance degradation (17.5 for SP and 19.4 or DP on macro-recall) shows that current models in their original form do not generalize well to unseen interactions. In contrast, we see that the performance gaps are smaller between the two splits for the hybrid models. And finally, we see clear empirical evidence that the MGAF module helps the model learn stronger representations for recognizing interactions from videos, outperforming all other models for both tasks. 

\section{Conclusion}
We presented an approach that utilizes the top-down structure implicit in a sequence of object detections to guide the video model to learn representation that captures dynamic aspects of complex human object interactions. We have shown that a bottom-up dual-pathway approach combined with the Motion Guided Attention Fusion module achieves this goal and leads to a video model that can even recognize humans interacting with previously unseen objects. We validate our approach on the Something-Else and IKEA-Assembly datasets where we achieve state of the art performance on recognizing compositional actions.
{\small
\bibliographystyle{ieee_fullname}
\bibliography{egbib}
}

\end{document}